\documentclass[letterpaper, 10 pt, conference]{ieeeconf}  

\IEEEoverridecommandlockouts                              

\usepackage{graphicx} 
\usepackage{color}
\usepackage{booktabs}
\usepackage{multirow} 
\usepackage{caption}
\usepackage{lipsum}
\usepackage{tabularx}
\usepackage{float}
\usepackage{stfloats}
\usepackage{makecell}
\usepackage{amsmath,amsfonts}
\usepackage{algorithmic}
\usepackage{url}
\usepackage{cite}
\usepackage[colorlinks=true, linkcolor=blue, citecolor=blue, urlcolor=blue]{hyperref}

\title{\LARGE \bf
UAV-VLN: End-to-End Vision Language guided Navigation for UAVs
}

\author{Pranav Saxena$^{1}$, Nishant Raghuvanshi$^{1}$ and Neena Goveas$^{1}$
\thanks{$^{1}$P. Saxena, N. Raghuvanshi, N. Goveas are with Birla Institute of Technology and Science Pilani, K.K Birla Goa Campus, Goa, India (e-mail: f20220257@goa.bits-pilani.ac.in; f20221052@goa.bits-pilani.ac.in; neena@goa.bits-pilani.ac.in).}%
}

\begin{document}

\maketitle
\thispagestyle{empty}
\pagestyle{empty}

\begin{abstract}

A core challenge in AI-guided autonomy is enabling agents to navigate realistically and effectively in previously unseen environments based on natural language commands. We propose UAV-VLN, a novel end-to-end Vision-Language Navigation (VLN) framework for Unmanned Aerial Vehicles (UAVs) that seamlessly integrates Large Language Models (LLMs) with visual perception to facilitate human-interactive navigation. Our system interprets free-form natural language instructions, grounds them into visual observations, and plans feasible aerial trajectories in diverse environments.

UAV-VLN leverages the common-sense reasoning capabilities of LLMs to parse high-level semantic goals, while a vision model detects and localizes semantically relevant objects in the environment. By fusing these modalities, the UAV can reason about spatial relationships, disambiguate references in human instructions, and plan context-aware behaviors with minimal task-specific supervision. To ensure robust and interpretable decision-making, the framework includes a cross-modal grounding mechanism that aligns linguistic intent with visual context.

We evaluate UAV-VLN across diverse indoor and outdoor navigation scenarios, demonstrating its ability to generalize to novel instructions and environments with minimal task-specific training. Our results show significant improvements in instruction-following accuracy and trajectory efficiency, highlighting the potential of LLM-driven vision-language interfaces for safe, intuitive, and generalizable UAV autonomy.

\end{abstract}

\section{INTRODUCTION}

Unmanned Aerial Vehicles (UAVs) are increasingly being integrated into diverse indoor and outdoor environments, performing a wide range of tasks such as package delivery~\cite{9129495,2110.02429}, aerial surveillance~\cite{1805.00881}, and search and rescue operations~\cite{10.1145/2750675.2750683,Quero2025}. These applications often require UAVs to operate in dynamic, human-centric settings, where they must navigate complex environments while interacting with both static objects and moving agents. Unlike ground robots, UAVs must account for three-dimensional motion, varying altitudes, and environmental factors such as wind disturbances and visibility constraints.

In such scenarios, effective UAV autonomy depends on robust perception, precise motion planning, and safe interaction with humans and other aerial or ground-based agents. While early methods largely relied on pre-defined flight paths or GPS-based waypoints~\cite{8409395,pulsgps}, recent advances in vision-based perception, learning-driven navigation, and onboard decision-making have significantly enhanced UAV adaptability and resilience. Nevertheless, real-world deployments continue to pose challenges due to dynamic environments, uncertainty, and incomplete information. Vision-language navigation (VLN) for UAVs offers a promising direction, enabling agents to interpret high-level natural language instructions and translate them into complex navigation tasks using visual inputs. By integrating multimodal reasoning, UAVs can enhance their situational awareness and responsiveness, leading to safer and more efficient autonomous operations in human-centric environments

\begin{figure}[t]
    \centering
    \includegraphics[width=0.45\textwidth]{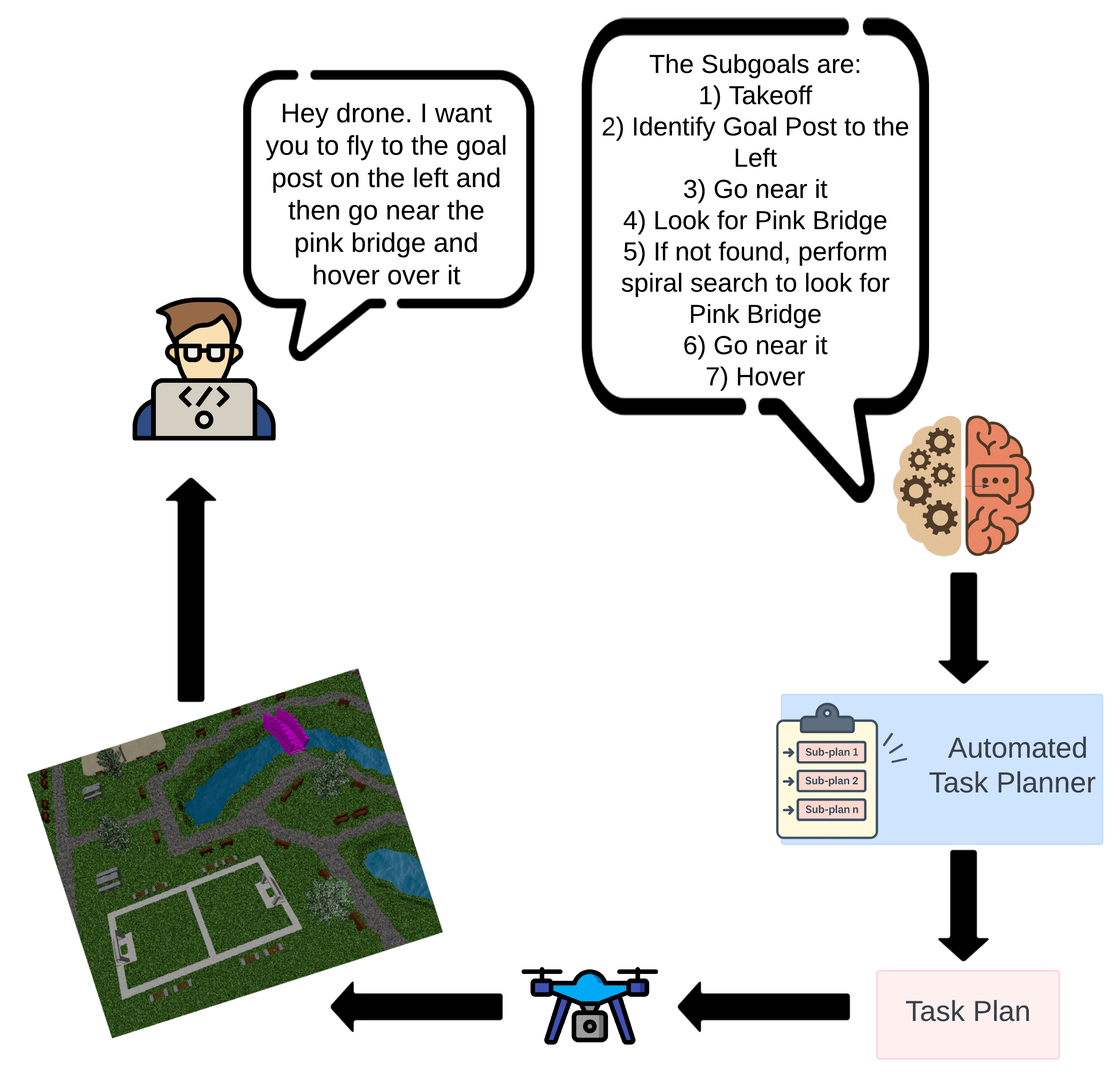}
    \caption{An example of VLN episode. A human user queries a LLM with the task prompt to generate sub-goals. An automated task planner generates
a task plan with respect to the sub-goals for the drone to execute. The drone also uses the visual input to carry out the task plan.\vspace{-1em}}
    \label{fig:intro}
\end{figure}

Recent advancements in Vision-Language Navigation (VLN) have significantly improved autonomous navigation for ground robots, enabling them to follow natural language instructions while using visual perception to make real-time decisions~\cite{exploreeqa2024, yu2023l3mvn,goetting2024endtoend,khanna2024goatbench}. Methods leveraging transformer-based architectures, imitation learning, and reinforcement learning have led to improvements in generalization across unseen environments and instruction grounding. Benchmark datasets such as Room-to-Room (R2R), Room-for-Room (R4R), and Touchdown have facilitated research in urban and indoor VLN settings, pushing the boundaries of embodied AI~\cite{mattersim,2010.07954,2406.09246,2109.08238}. However, these methods are primarily designed for wheeled or legged robots, which operate in structured 2D environments with well-defined constraints.

In contrast, there is limited exploration of VLN in the context of Unmanned Aerial Vehicles (UAVs)—despite their increasing deployment in real-world outdoor applications. UAV navigation presents unique challenges, including reasoning over unstructured 3D environments, managing altitude variations, and adapting to dynamic conditions such as weather and terrain. These complexities demand new paradigms that move beyond planar navigation, requiring aerial-specific perception, adaptive flight planning, and real-time decision-making.

To address this gap, we introduce UAV-VLN, a novel end-to-end framework for vision-and-language-based navigation tailored to UAV platforms. Our approach leverages a fine-tuned Large Language Model (LLM) to interpret natural language instructions, which are grounded in visual context to guide the UAV through complex environments, as shown in Figure~\ref{fig:intro}.

Our key contributions are:

(i) We propose UAV-VLN, a new paradigm for human-interpretable vision-language navigation for aerial robots, enabling natural instruction following in real-world UAV settings.

(ii) We construct a novel dataset comprising over 1,000 aerial navigation instruction prompts and corresponding sub-plans, designed to train and evaluate large-language models in 3D UAV contexts.

(iii) We demonstrate that our method generalizes to unseen environments and instructions, achieving robust zero-shot navigation performance in both indoor and outdoor settings

\section{Related Work}

\subsection{Vision-and-Language Navigation}
Vision-and-Language Navigation (VLN) is a multimodal task that involves interpreting natural language instructions and visual observations to predict navigation actions. Early VLN research focused primarily on indoor, graph-constrained environments. Datasets such as Room-to-Room and REVERIE, built on Matterport3D, employed discrete navigation graphs to simulate agent movement through household scenes. To address the limitations of graph-based navigation, VLN-CE~\cite{krantz_vlnce_2020} introduced continuous control in 3D reconstructed environments, enabling more realistic agent dynamics.

Other VLN efforts have explored interactive or dialog-based instruction-following~\cite{nguyen2018vision, thomason:corl19} or synthetic household environments like ALFRED, TEACh~\cite{ALFRED20,teach}, where agents follow high-level instructions to complete embodied tasks.

Beyond indoor environments, datasets such as Touchdown~\cite{1811.12354} and LANI extended VLN to outdoor urban scenes, though these are often based on street-level imagery and limited agent mobility. Drone-based navigation was explored in Modified LANI~\cite{1809.00786}, but their simplified simulation environment and lack of altitude control limit realism and transferability to real-world aerial settings.

In contrast, UAV-VLN builds on these foundations by extending vision-language navigation to unstructured indoor and outdoor environments using UAVs with full 3D control. Our framework leverages grounded vision and language understanding to operate in complex, dynamic spaces without relying on discretized action spaces or heavily simulated scenes. This positions UAV-VLN as a more realistic and scalable solution for instruction-driven aerial autonomy

\begin{figure*}[t]
    \centering
    \includegraphics[width=0.9\textwidth]{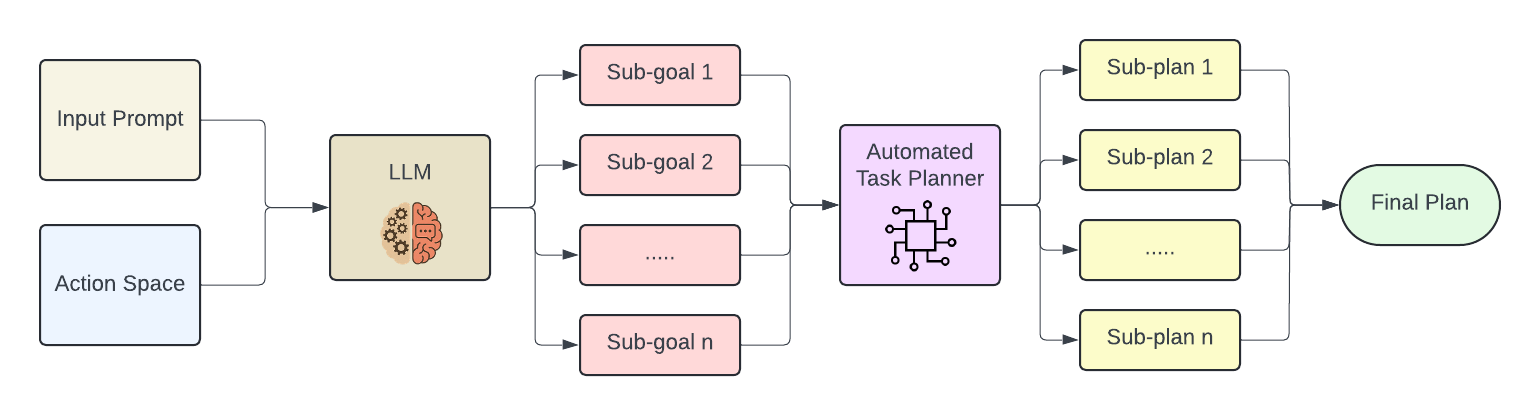}
    \caption{The system architecture of UAV-VLN with four sequential stages: (1) Natural language prompt and action space are provided as inputs; (2) a fine-tuned LLM performs semantic decomposition of the instruction into structured sub-goals; (3) an automated task planner maps each sub-goal to executable low-level UAV actions considering environmental context; (4) the generated sub-plans are synthesized into a coherent final mission plan ensuring instruction consistency, safety, and robustness.}.\vspace{-2em}
    \label{fig:arch}
\end{figure*}

\subsection{Aerial Navigation}
While vision-only aerial navigation has been actively explored in the robotics community, the integration of both vision and language for UAV navigation remains relatively underexplored. Prior works~\cite{8264734,Majdik2017-fy} have leveraged pre-collected real-world drone datasets to address perception-driven control and navigation. However, collecting diverse aerial data is resource-intensive and poses safety risks, leading many studies to adopt simulation environments, which offer scalable training with rich ground-truth annotations.

Despite these advancements, most existing approaches overlook the language modality, limiting task complexity and the expressiveness of goal specification. In contrast, our work leverages natural language as the primary driver of navigation, enabling UAVs to interpret nuanced instructions and exhibit context-aware, goal-directed behavior. This multimodal formulation facilitates more expressive, flexible, and human-aligned interaction paradigms for real-world UAV autonomy.
\subsection{Foundation Models for Embodied Navigation}

Recent advancements in large-scale foundation models, such as Large Language Models (LLMs) and Vision-Language Models (VLMs), have demonstrated immense potential in robotic navigation~\cite{google2024gemini2flash,2407.21783,2307.09288,2303.08774,2310.06825}. SayCan~\cite{2204.01691} integrates LLMs for high-level task planning by grounding textual instructions in robotic affordances. GPT-Driver~\cite{2310.01415} frames motion planning as a language modeling task, evaluating GPT-3.5 in autonomous driving simulations. L3MVN~\cite{2304.05501} builds semantic representations of environments and employs LLMs to achieve long-horizon navigation goals, while LLaDA~\cite{2502.09992} adapts driving behavior to diverse regional traffic rules using LLM-guided reasoning. LM-Nav~\cite{2207.04429} combines GPT-3 with CLIP to navigate outdoor scenes from natural language instructions, leveraging both linguistic and visual cues for effective path planning. These works showcase the growing ability of foundation models to generalize across domains and modalities, yet few have explored their integration into UAV platforms for grounded, instruction-following navigation in realistic 3D environments

\section{OUR APPROACH}

We present UAV-VLN, a novel end-to-end vision-language navigation framework that enables Unmanned Aerial Vehicles (UAVs) to interpret and execute free-form natural language instructions in complex, real-world environments. UAV-VLN is designed to bridge the gap between high-level human intent and low-level aerial control by leveraging the complementary strengths of large language models (LLMs) and visual perception systems.

Unlike existing VLN approaches which predominantly focus on ground robots in structured indoor settings, UAV-VLN is tailored for the challenges of 3D aerial navigation, including dynamic obstacle avoidance, variable altitude control, and spatial grounding in visually complex scenes. Our framework integrates a fine-tuned LLM for semantic instruction parsing, a vision model for real-time scene understanding, and a cross-modal grounding module that aligns linguistic goals with visual context to guide UAV behavior. This integration enables the UAV to reason over spatial relations, disambiguate object references, and plan trajectories that satisfy the instruction semantics with minimal supervision.

\subsection{Problem Definition}

The \textbf{Vision-Language Navigation (VLN)} task for UAVs can be defined as follows:\\
Given a free-form natural language instruction \( I \) and a visual observation stream \( V = \{v_1, v_2, ..., v_T\} \) captured from the UAV's onboard RGB camera, the goal is to predict a sequence of control commands \( A = \{a_1, a_2, ..., a_T\} \) that guides the UAV from its starting position to the target location or goal state described in \( I \), while navigating safely through the environment.

Key challenges include:
\begin{itemize}
    \item \textbf{Semantic Parsing}: Extracting actionable goals and spatial cues from unstructured language.
    \item \textbf{Visual Grounding}: Aligning language-referenced objects and regions with the UAV’s visual field in dynamic, unstructured environments.
    \item \textbf{Trajectory Planning}: Generating feasible, safe, and instruction-consistent flight paths in 3D space.
    \item \textbf{Generalization}: Maintaining robustness across novel environments, instructions, and visual scenes with minimal retraining.
\end{itemize}

Our approach addresses this task through a modular pipeline that combines LLM-driven intent interpretation, visual-semantic alignment, and task-aware planning, forming a unified control policy for instruction-following UAV navigation.

\subsection{Natural Language Prompting}

A crucial component of Vision-and-Language Navigation (VLN) is the system's ability to accurately interpret and act upon natural language instructions. While prior work—especially in ground robot navigation—has successfully utilized general-purpose, pre-trained large language models (LLMs) like \textit{ChatGPT} or \textit{Gemini}, we found these models occasionally misinterpret or misclassify actions in UAV navigation. As these models are online and reliant on cloud infrastructure, they can also suffer from latency or availability issues. Given the high sensitivity and safety demands of UAVs, such errors may lead to catastrophic failures, including crashes or unintended behavior.

To ensure higher reliability and precision in instruction following, we adopt a fine-tuning approach based on a domain-specific dataset. We curate a custom UAV instruction dataset and fine-tune the \textbf{TinyLlama-1.1B}~\cite{zhang2024tinyllama} model on this data. This targeted adaptation allows the language model to better understand UAV-specific terminology, spatial instructions, and safety-critical nuances. In practice, we observe a substantial improvement in the accuracy and consistency of responses, significantly lowering the risk of erroneous actions and improving the overall robustness of our navigation pipeline.

As shown in Figure~\ref{fig:arch}, the fine-tuned LLM takes the following two key inputs:
\begin{enumerate}
    \item \textbf{Input Prompt:} A high-level natural language instruction from the user.
    \item \textbf{Action Space:} The set of all valid discrete actions the UAV can perform.
\end{enumerate}

The LLM then generates a sequence of intermediate \textit{sub-goals}, each corresponding to an executable UAV action. This structured output enables safe, interpretable, and step-by-step execution of complex instructions in dynamic environments.


\subsection{Automated Task Planner}

Once the fine-tuned LLM decomposes the natural language instruction into a sequence of interpretable sub-goals, these sub-goals must be further converted into concrete action plans that can be executed by the UAV in the physical environment. To facilitate this translation, we introduce an \textit{Automated Task Planner} that maps each high-level sub-goal to a detailed sequence of low-level control commands.

The task planner leverages the discrete action space of the UAV and considers the current state and environment context to generate valid and efficient sub-plans for each sub-goal. These sub-plans are then composed into a coherent \textbf{final execution plan}, ensuring the UAV performs the task safely and optimally.

We implement our control pipeline using the \textbf{Robot Operating System 2 (ROS 2)}, which provides modularity, real-time capabilities, and robust integration with UAV flight stacks. ROS 2 nodes handle execution, sensor feedback, and safety checks, allowing smooth integration between high-level planning and low-level control.

As shown in Figure~\ref{fig:arch}, the planner bridges the gap between abstract language-derived sub-goals and executable UAV trajectories, ensuring real-world feasibility and safety.

\subsection{Visual Input}

Analyzing visual input in conjunction with language understanding is a key factor in determining the target position of the UAV. In the context of Vision-and-Language Navigation (VLN), it is crucial that the UAV not only perceives the environment accurately but also grounds its perception based on natural language instructions.

Recent advances in open-vocabulary detection and segmentation \cite{2303.05499, 2112.10003,ravi2024sam2} have enabled systems to localize objects and regions in an image based on free-form text. While some methods rely on closed-vocabulary models \cite{1506.02640, 1512.02325}, we adopt an open-vocabulary approach to enhance generalization and flexibility across diverse scenes.

To this end, we utilize \textbf{Grounding DINO}, a transformer-based open-vocabulary object detector. It effectively leverages the semantic richness of textual queries to localize relevant entities in the visual input. Given the instruction and the processed text from our fine-tuned TinyLlama-1.1B model, we extract relevant object or region descriptors that are then used as queries for Grounding DINO.

This allows the UAV to:
\begin{enumerate}
    \item Interpret the instruction to identify target objects or landmarks.
    \item Use Grounding DINO to localize those targets in the camera feed.
    \item Generate grounded sub-goals based on the spatial relationship between the UAV and the detected entities.
\end{enumerate}

This visual-linguistic grounding pipeline enhances the UAV's ability to understand ambiguous or context-dependent instructions and translate them into precise navigation targets. It also enables our system to operate in open-world settings without requiring predefined object vocabularies.

\subsection{Termination}

In Vision-and-Language Navigation (VLN), accurately determining when to terminate the navigation task is as critical as executing the path itself. An early or late termination can lead to undesired behaviors such as hovering indefinitely, drifting, or missing the intended goal location.

Our approach defines a robust \textbf{termination criterion} that combines both \textit{visual confirmation} and \textit{language-grounded cues}. Specifically, once the UAV executes the planned set of sub-goals and reaches the vicinity of the final target location, it uses the following checks:

\begin{enumerate}
    \item \textbf{Goal Object Detection:} Using Grounding DINO, the UAV confirms the presence of the instructed target object or landmark in its current view.
    \item \textbf{Proximity Check:} A predefined spatial threshold is used to verify whether the UAV is within an acceptable range of the target, based on real-world coordinates or estimated visual alignment.
    \item \textbf{Instruction Satisfaction:} The UAV verifies if the sub-goals derived from the instruction have been successfully executed.
\end{enumerate}

Only when all of these criteria are satisfied does the UAV initiate a safe hover or land routine to complete the mission. This termination logic is implemented in the ROS2 control stack to ensure safe handling of the UAV state during task conclusion. This integration also allows for future extensions such as user-initiated stop signals or dynamic re-planning in case of mission failure.

\begin{figure}[t]
    \centering
    \includegraphics[width=0.47\textwidth]{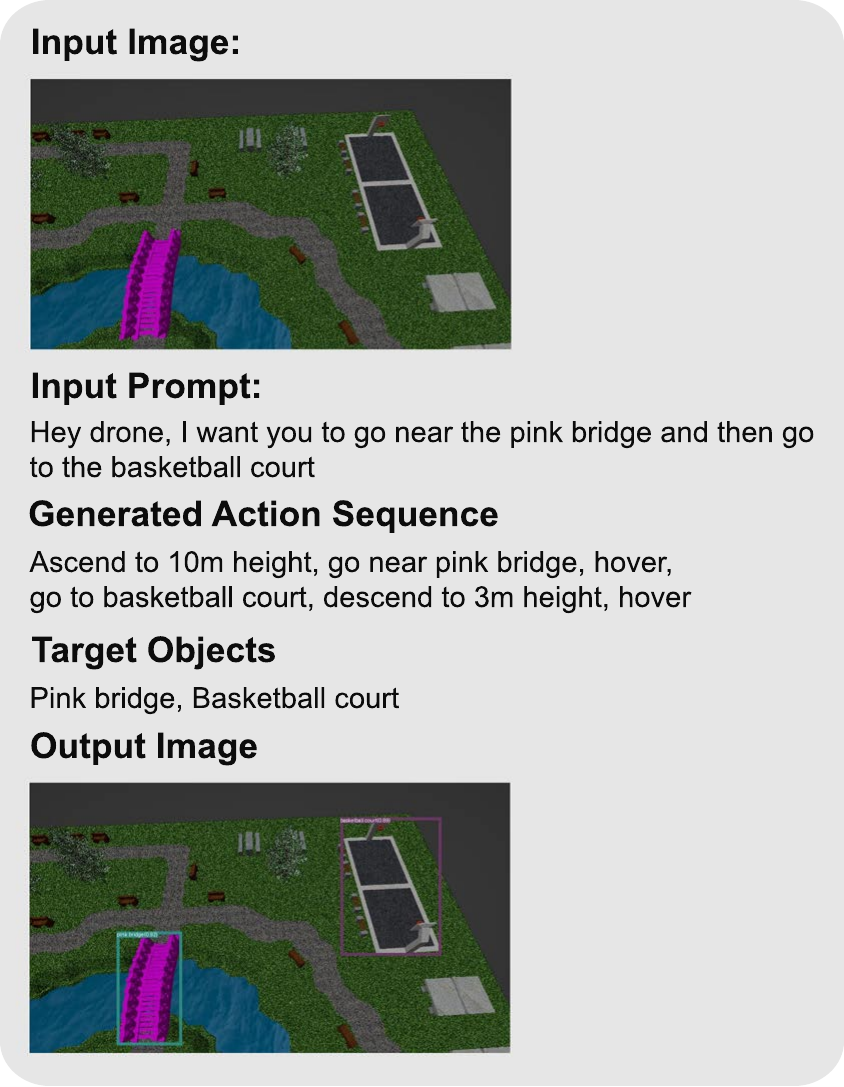}
    \caption{An example of an input image and natural language prompt used in UAV-VLN.\\(i) The prompt is processed by TinyLLaMA to generate a high-level action sequence.\\(ii) An automated task planner identifies the relevant target objects in the sequence.\\(iii) The input image and identified objects are passed to Grounding-DINO to obtain bounding boxes.\\(iv) The action sequence is then mapped to low-level drone commands for execution.\vspace{-1.47em}}
    \label{fig:expt}
\end{figure}

\section{EXPERIMENTS}

We evaluate our approach on four scenes with different objects, performing 15 tasks per scene and recording the results. We also analyze how various design parameters—such as using non-finetuned LLMs and the choice of open or closed-vocabulary detection/segmentation models—affect the performance of the end-to-end Vision-Language Navigation (VLN) framework.

\subsection{Procedure}

\textbf{Experimental Setup:} Our experiments are run on an Nvidia GTX 1650 GPU integrated into a laptop. This setup demonstrates that our framework is capable of operating on a drone equipped with a capable companion computer and GPU.

\textbf{Simulator:} We leverage Gazebo Garden with Robot Operating System 2 (ROS2) for simulations. The drone in our simulations is equipped with a Pixhawk Flight Controller and a monocular camera mounted on the bottom.

\textbf{Metrics:} Following common practice in VLN research, we use the following evaluation metrics:

\begin{enumerate}
    \item \textbf{Success Rate (SR)}: The fraction of episodes that are successfully completed.
    \item \textbf{Success Rate Weighted by Inverse Path Length (SPL)}: A measure of path efficiency that penalizes longer paths.
\end{enumerate}

\subsection{Evaluation}

\begin{table*}[!t]
\small
\center
\setlength\tabcolsep{9pt}
\renewcommand{\arraystretch}{1.6}
\captionof{table}{Quantitative Results — Performance Comparison of DEPS, VLMNav, and UAV-VLN (Ours)}
\label{tab:scene_table}

\begin{tabularx}{\textwidth}{ll|cc|cc|cc|cc}  
\hline
& & \multicolumn{2}{c|}{\textbf{Warehouse}} & \multicolumn{2}{c|}{\textbf{Park}} & \multicolumn{2}{c|}{\textbf{House Neighborhood}} & \multicolumn{2}{c}{\textbf{Office}}\\ 
\hline 
& \textbf{} & \textbf{SR} & \textbf{SPL} & \textbf{SR} & \textbf{SPL} & \textbf{SR} & \textbf{SPL} & \textbf{SR} & \textbf{SPL} \\ 
\hline  

\parbox[c]{1.2mm}{\multirow{3}{*}}
& DEPS & 73.33\% & 0.1704 & 86.67\% & 0.0733 & 80\% & 0.2875 & 80\% & 0.8284  \\ 
& VLMNav & 80\% & 0.1833 & 73.33\% & 0.0755 & 66.67\% & 0.2519 & \textbf{86.67\%} & \textbf{0.8667}  \\  
& Ours & \textbf{86.67\%} & \textbf{0.2070} & \textbf{93.33\%} & \textbf{0.0792} & \textbf{86.67\%} & \textbf{0.3014} & \textbf{86.67\%} & 0.795 \\  
\hline  

\end{tabularx}
\vspace{-5pt}
\end{table*}

We evaluate our framework across four scenes, each with 15 episodes featuring different navigation prompts. For each scene, we compute the success rate to quantify the effectiveness of our approach.

\begin{itemize}
    \item Scene-1: Warehouse
    \item Scene-2: Park
    \item Scene-3: House Neighborhood
    \item Scene-4: Office
\end{itemize}

\subsection{Baselines}
We select two popular VLN approaches, which we have adapted for UAVs. We then compare our approach against these adapted methods:

(i) \textbf{DEPS: Describe, Explain, Plan and Select}~\cite{2302.01560}: It uses LLMs to perform intermediate reasoning by describing the environment, explaining subgoals, planning candidate actions, and selecting feasible plans grounded in visual observations.

(ii) \textbf{VLMNav}~\cite{goetting2024endtoend}: It uses Gemini 2.0 Flash as a zero-shot and end-to-end language conditioned navigation policy. 

\subsection{Ablation Study}

In this section, we conduct an ablation study to compare the effects of different design choices on the performance of our framework. Specifically, we evaluate variations in the LLM architecture and the vision model used in the framework.

\begin{table}[H]
    \centering
    \renewcommand{\arraystretch}{0.95}
    \begin{tabular}{l *{4}{c}}
        \toprule
        Design Choices & Warehouse &Park &House &Office \\
        \midrule
        Gemini+YOLO    & 26.67 & 13.33 & 6.67 & 26.67 \\
        TinyLlama+YOLO              & 20 & 13.33 & 6.67 & 26.67 \\
        Gemini+ClipSeg           & 73.33 & 80 & 73.33 & 80 \\
        TinyLlama+ClipSeg                    & 66.67 & 73.33 & 66.67 & 73.33 \\
        Gemini+Grounding-DINO           & 80 & 86.67 & 80 & \textbf{86.67} \\
        TinyLlama+Grounding-DINO                    & 73.33 & 80 & 80 & 80 \\

        \midrule
        UAV-VLN (YOLO)  & 33.33 & 13.33  & 6.67 & 33.33 \\
        UAV-VLN (ClipSeg)      & 73.33 & 86.67 & 80 & 80 \\
        UAV-VLN (Grounding-DINO)           & \textbf{86.67} & \textbf{93.33} & \textbf{86.67} & \textbf{86.67} \\
        \bottomrule
    \end{tabular}
    \caption{Success rates $[\%]$ of different design choices. The results for each scene are averaged across all episodes. The upper part of the table shows the results of the baselines based on different LLM and Vision Models. The lower part lists the outcome of UAV-VLN with fine-tuned \textit{with TinyLlama-1.1B} and different vision models.}
    \label{tab:sr}
\end{table}

\section{CONCLUSION}

In this work, we present UAV-VLN, a novel end-to-end Vision-Language Navigation framework for UAVs that reframes aerial navigation as a question-answering task. By combining the semantic reasoning capabilities of fine-tuned large language models with open-vocabulary visual grounding, our system enables UAVs to interpret free-form instructions and navigate complex, dynamic environments with minimal task supervision.

Through comprehensive evaluations across diverse aerial scenes and an ablation study comparing key design choices, we show that UAV-VLN significantly improves instruction-following accuracy and path efficiency over baseline methods. Our ablations reveal that closed-vocabulary object detectors like YOLO underperform significantly, while open-vocabulary models such as CLIPSeg and Grounding DINO offer better generalization and robust grounding. Furthermore, our fine-tuned TinyLLaMA outperformed both the base TinyLLaMA and Gemini models, highlighting the importance of tailored language model adaptation for UAV tasks.

In future work, we aim to incorporate navigation history and lightweight semantic mapping to help UAVs reason globally, avoid redundant exploration, and plan more efficient paths. This would move UAV-VLN systems closer to truly
scalable and lifelong navigation in challenging open-world environments.
\\

\bibliographystyle{unsrt}
\bibliography{references}

\end{document}